\documentclass[sigconf]{acmart}
\usepackage{pifont}% http://ctan.org/pkg/pifont
\usepackage{multirow}
\AtBeginDocument{%
  }

\setcopyright{acmlicensed}
\copyrightyear{2024}
\acmYear{2024}
\acmConference[Conference acronym 'XX]{Make sure to enter the correct
  conference title from your rights confirmation emai}{
  2024}{}  

\begin{document}

\title{Low-Rank Adaptation of Time Series Foundational Models for Out-of-Domain Modality Forecasting}

\author{Divij Gupta, Anubhav Bhatti, Suraj Parmar, Chen Dan, Yuwei Liu, Bingjie Shen, San Lee}
\affiliation{%
  \institution{AI Engineering Team, SpassMed Inc.}
  \city{Toronto}
  \state{Ontario}
  \country{Canada}
}

\begin{abstract}

Low-Rank Adaptation (LoRA) is a widely used technique for fine-tuning large pre-trained or foundational models across different modalities and tasks. However, its application to time series data, particularly within foundational models, remains underexplored. This paper examines the impact of LoRA on contemporary time series foundational models: Lag-Llama, MOIRAI, and Chronos. We demonstrate LoRA's fine-tuning potential for forecasting the vital signs of sepsis patients in intensive care units (ICUs), emphasizing the models' adaptability to previously unseen, out-of-domain modalities. Integrating LoRA aims to enhance forecasting performance while reducing inefficiencies associated with fine-tuning large models on limited domain-specific data. Our experiments show that LoRA fine-tuning of time series foundational models significantly improves forecasting, achieving results comparable to state-of-the-art models trained from scratch on similar modalities. We conduct comprehensive ablation studies to demonstrate the trade-offs between the number of tunable parameters and forecasting performance and assess the impact of varying LoRA matrix ranks on model performance.

\end{abstract}



\maketitle

\section{Introduction}

Large pre-trained or foundational models (FMs) have become pivotal in deep learning advancements, providing robust backbone architectures \cite{foundation}. They offer rich, general-purpose feature representations across various domains like language \cite{text_survey} and vision \cite{vision_survey} through comprehensive self-supervised pre-training on vast unlabeled datasets. While they can be deployed in zero-shot or fine-tuned settings, full model weight modification risks reducing generalizability and creating inefficiencies. Parameter-Efficient Fine-Tuning (PEFT) algorithms address this by fine-tuning a minimal set of tailored weights instead of adjusting the entire model. Low-Rank Adaptation (LoRA) \cite{lora}, a popular PEFT technique, is known for its simplicity and effectiveness. Although PEFT techniques like LoRA have been successfully applied across domains \cite{peft_survey}, their use in time series FMs (TSFMs) is still underexplored, given the recent surge of interest in this area.

\begin{figure*}
  \centering
  \includegraphics[width=0.85\linewidth]{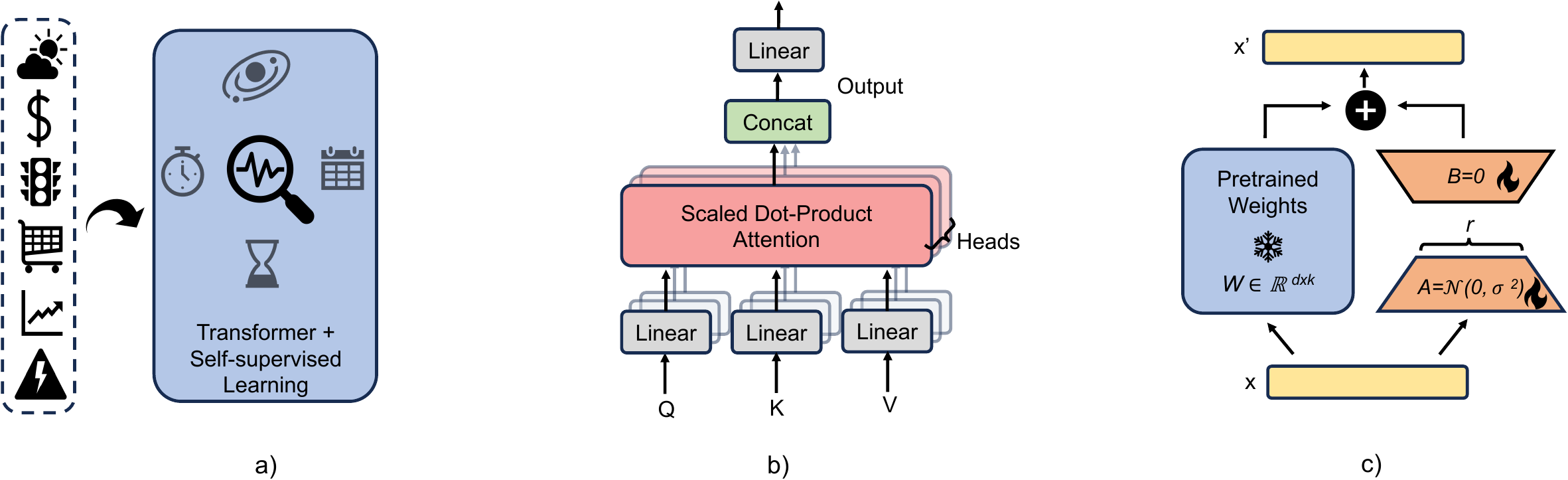}
  \caption{The primary components of our method: a) A transformer-based TSFM for learning rich feature representations from time series data; b) Multi-Head Self-Attention \cite{attention}; c) The Low-Rank Adaptation (LoRA) mechanism.}
  \label{overall}
\end{figure*}

Time series data, like text, encompass various domains, including finance, weather, energy, retail, and others. However, unlike text, building FMs for time series presents numerous challenges, such as varying observation frequencies, multiple varieties, and non-standard quantization of values. Researchers have recently addressed some of these issues \cite{chronos, moirai}. The challenge is particularly acute in domains like healthcare, where data is both sensitive and scarce. Considerable research has been conducted on utilizing time series data for health and well-being analysis through relatively smaller, data-specific methods. These include activity recognition \cite{har}, emotion detection \cite{emotion, anubhav_emotion}, cognitive load assessment \cite{prithila_load}, remote health monitoring \cite{divij_iot}, stress detection \cite{anubhav_stress}, sepsis onset prediction \cite{sepsis}, vital signs forecasting \cite{spass1}, and others. However, the application of domain-adapted FMs to time series data in healthcare remains largely unexplored. To this end, in this paper, we employ FMs specifically designed for time series to forecast vital signs in sepsis patients in intensive care units (ICUs). We adapt these models using LoRA and thoroughly analyze the results. 

To summarize, our contributions are as follows: \textbf{1.} We explore the efficiency and impact of LoRA-based fine-tuning across three leading TSFM architectures: Lag-Llama \cite{lagllama}, MOIRAI \cite{moirai}, and Chronos \cite{chronos} for vital sign forecasting on the publicly available eICU Collaborative Research Database \cite{eicu}.
\textbf{2.} We conduct a thorough comparative analysis of various fine-tuning methodologies, including zero-shot, full fine-tuning, and LoRA-based selective fine-tuning. Our analysis reveals that LoRA fine-tuning improves forecasting performance while significantly reducing computational requirements compared to full fine-tuning. 
\textbf{3.} We compare our results with the state-of-the-art (SOTA) approach in vital sign forecasting and illustrate that LoRA fine-tuning enables certain Chronos variants to match or surpass SOTA performance. This underscores the applicability and efficiency of LoRA fine-tuning in practical forecasting tasks.
\textbf{4.} We conducted thorough ablation studies to examine the trade-off between performance and the number of trainable parameters. Additionally, we assessed the impact of varying rank ($r$) of LoRA matrices on the model performance.

\vspace{-0.1 cm}

\section{Related Work}
\label{tsfm}

\noindent{\textbf{Parameter-Efficient Fine-Tuning (PEFT) Techniques}}: 
PEFT techniques can be broadly categorized into two groups: selective and additive approaches. Selective PEFT involves fine-tuning a select type of parameters in the FMs, which help it adapt to domain knowledge. For instance, in \cite{attn_tune}, the authors fine-tune the attention layers of the transformer architecture \cite{attention} in the FM. Similarly, studies \cite{bitfit} and \cite{layer_norm} fine-tune only the bias terms and the layer-normalization (LayerNorm) layers of the transformer architecture, respectively. On the contrary, additive PEFT introduces a limited set of new weights into the FM and only these new weights, also known as adapter modules, are fine-tuned. Similar to selective PEFT techniques, different adapter modules \cite{adapterh, adapterl, adapterd, lora} offer distinct strategies for fine-tuning the FM. Among them, LoRA \cite{lora} has demonstrated superior performance on baseline tasks while maintaining a minimal number of trainable parameters, leading to its widespread adoption across diverse fields \cite{peft_survey, midl_peft, lora_use_vlm}. However, its efficacy in fine-tuning TSFMs remains unexplored and is the central focus of this study. Information on incorporating LoRA with the transformer-based architecture is mentioned in the Section \ref{lora_tech}. \\

\noindent{\textbf{Time Series Foundational Models (TFSM)}}: TimeGPT \cite{timegpt} is a pioneering TSFM that uses an encoder-decoder transformer-based architecture for time series forecasting. However, due to its proprietary nature, limited information about the model is available publicly.
Lag-Llama \cite{lagllama} employs a decoder-only transformer architecture, drawing inspiration from the LLaMA \cite{llama} large language model (LLM). This model utilizes the LLM architecture for time series forecasting by using individual observations, lagged features from previous observations, and timestamps to capture frequency information. MOIRAI \cite{moirai}, utilizing an encoder-only transformer architecture, processes time series data in non-overlapping patches rather than at the individual datapoint level. This approach provides enhanced contextual information for forecasting subsequent patches. Patch sizes vary based on the data frequency, with smaller patches used for low-frequency data and larger patches for high-frequency data. Chronos \cite{chronos} employs the encoder-decoder transformer architecture from the T5 \cite{t5} LLM family to perform time series forecasting. The model simplifies data processing by discretizing real-valued inputs through a binning and scaling function. Unlike other models, Chronos does not include specialized mechanisms for handling time-frequency information. Although studies like \cite{gpt4mts, llmtime} adapt or fine-tune LLMs for time series forecasting, this research focuses exclusively on models trained \textit{from scratch} on time series data. Thus, only open-source TSFMs demonstrating SOTA performance are considered \cite{chronos}.

\section{Method}
\label{lora_tech}

This section discusses the method we used in our study's experiments. First, we clearly define our problem statement, the backbone architecture we use to predict the required values, and how we can provide additional support to the backbone architecture through domain adaptation. \\

\noindent{\textbf{Problem Statement}}:
Let $D$ be a dataset consisting of multiple vital signs in the form of time series such that any time series, $t$, can be represented as $t = {t_1, t_2, ..., t_T}$ where $t_i, (i \in {1,2,..., T})$ represents the value at the $i^{th}$ time-step for a total of $T$ time-steps. Given a context length $C$ $(1<C<T)$, we want to use the time series $t_{1:C} = {t_1, t_2,..., t_C}$ to forecast the time series $t_{C:C+h} = {t_{C+1}, t_{C+2},..., t_{C+h}}$, where $h$ is the forecast horizon, and $C+h=T$. \\

\noindent{\textbf{Backbone}}:
In addressing the forecasting task outlined above, we employ various TSFMs that have been pre-trained to forecast across a broad spectrum of time series data. 
Throughout pre-training, these models gain insights into the underlying structure and complexities of the temporal domain, which we harness to forecast $t_{C:C+h}$.
For TSFM $M$, with weight parameters $\theta$, the objective is achieved by: 
\begin{equation}   
    M_{\theta} ({t_1, t_2,..., t_C}) = t_{C+1}, t_{C+2},..., t_{C+h}.
\end{equation}
For our method, we use three choices for $M$, namely, Lag-Llama \cite{lagllama}, MOIRAI \cite{moirai}, and Chronos \cite{chronos}, which have been briefly discussed in Section \ref{tsfm}. These FMs have different transformer-based structures, pre-training and data processing schemes, and number of parameters, among other distinguishing factors. 
Since Chronos and Lag-Llama perform univariate forecasts, the same setting for MOIRAI has been followed for a fair comparison. Thereby, for all three candidates for our backbone, we implement the univariate setting wherein the model takes in univariate input, without any covariate, except the timestamp, and forecasts univariate outputs. 
It is crucial to emphasize that, despite being pre-trained on time series data across multiple domains, these models lack domain-specific knowledge of vital signs' time series data. This characteristic makes them more adaptable and provides additional flexibility to optimize their performance for the specific task. \\

\noindent{\textbf{Parameter-Efficient Domain Adaptation}}:
We use LoRA to provide domain knowledge to $M$ by fine-tuning a minimal number of weights. As described earlier, LoRA introduces small additive weights in existing linear weight matrices. When multiplied, the weight matrices are the same shape as the original weight matrix and provide a parallel trainable pathway to the incoming feature map. 
Fundamentally, the concept underlying this approach is that during adaptation, updates to the weight matrices exhibit low intrinsic ranks \cite{lora}. Specifically, the modification is performed by adding a rank-constrained product of matrices $A$ and $B$. For a weight matrix $W \in \mathbb{R}^{d \times k}$, the update is represented as $\Delta W = BA$, where $B \in \mathbb{R}^{d \times r}$ and $A \in \mathbb{R}^{r \times k}$, and $r << min(d,k)$. Here, $d$ and $k$ denote the working dimensions of $W$, and $r$ denotes the rank of the adapters. During fine-tuning, $W$ is frozen, while the weights of $A$ and $B$ are updated. As given in \cite{lora}, we initialize $B$ with zeros and $A$ with a small set of random values sampled from a Gaussian distribution. The updated matrix, $W'$ is given by:

\begin{equation}
    W' = W + \frac{\alpha}{r} BA.
\end{equation}

In this formulation, $ \alpha $ is a scaling parameter that adjusts the influence of the update on the original weights. 
We adopt the four weight matrices, $W_q, W_k, W_v, W_o$, corresponding to the weight matrices of the Query (Q), Key (K), Value (V), and Output of the Multi-Head Self-Attention module (depicted in Figure \ref{overall}) of the transformer architecture used in $M$ to maximize the utilization of the attention mechanism, a setting used in \cite{lora}.
Thus, with the addition of LoRA, the required forecast can be predicted as:

\begin{equation}
    t_{C+1}, t_{C+2}, ..., t_{C+h} = M_{\theta '} ({t_1, t_2, ..., t_C}), 
\end{equation}

where $\theta ' = \theta + \phi$ where $\theta$ are the original frozen parameters of $M$, and $\phi$ is the additive parameters of the model introduced through LoRA. $\phi$ is learnt while fine-tuning on the vital signs data.

\begin{table}[t]
\caption{Results comparing MeanBP and HR forecasts across different TSFM settings: zero-shot, full fine-tuning, and LoRA fine-tuning ($r$=2, $\alpha$=16). The best values for each individual model are underlined, while the overall best values are highlighted in bold.}
\scriptsize

\begin{tabular}{c|c|ccc|ccc}
\hline

\multirow{2}{*}{Model}  & \multirow{2}{*} {Setting} & \multicolumn{3}{c}{MeanBP*}  & \multicolumn{3}{c}{HeartRate*}  \\
& & MSE $\downarrow$   & DTW  $\downarrow$  & MAPE  $\downarrow$ & MSE $\downarrow$  & DTW  $\downarrow$  & MAPE $\downarrow$  \\

\hline
\hline

\multirow{3}{*}{\begin{tabular}[c]{@{}c@{}}Bhatti et al.\\ \cite{spass1}\end{tabular}} 
    & N-HiTS    & 19.81  & \textbf{16.32}  & 7.90  & \underline{7.18}  & \textbf{7.92} & \underline{6.27}  \\
   & N-BEATS       & 27.42  & 18.60  & 9.64  & 12.98  & 17.90  & 9.15  \\
   & TFT       & \textbf{19.00}  & 23.46  & \textbf{7.78} & 7.57  & 15.79 & 6.52 \\  \hline
   
\multirow{3}{*}{Lag-Llama}
    & 0-shot    & 30.94  & 24.64  & 10.11  & 15.55 & 18.42  & 9.41 \\
   & Full FT       & \underline{22.93}    & \underline{19.25}   & \underline{8.23}  & 16.60     & 20.45   &  9.65 \\
    & LoRA FT       & 23.83  & 19.54  & 8.38  & \underline{10.39} & \underline{13.45}  & \underline{7.22}  \\ \hline

\multirow{3}{*}{\begin{tabular}[c]{@{}c@{}}Moirai\\ (Small)\end{tabular}}  
    & 0-shot    & 28.16  & \underline{19.18}  & 9.75 & 10.83 & 12.61  & 7.89 \\
   & Full FT       & \underline{21.84} & 20.95 & \underline{8.56} & \underline{9.49} & \underline{11.87} & \underline{7.30}  \\
   & LoRA FT       & 23.84  & 19.71  & 9.01  & 10.46  & 12.87 & 7.78  \\ \hline
   
\multirow{3}{*}{\begin{tabular}[c]{@{}c@{}}Moirai\\ (Base)\end{tabular}} 
    & 0-shot    & 28.90 & \underline{19.60}  & 9.86  & \underline{11.01}  & \underline{13.30}  & \underline{7.95} \\
   & Full FT       & \underline{25.46}  & 20.50 & \underline{9.39}  & 11.27 & 14.67  & 8.05  \\
   & LoRA FT       & 26.67  & 21.38  & 9.61  & 12.47 & 14.91  & 8.50  \\ \hline
   
\multirow{3}{*}{\begin{tabular}[c]{@{}c@{}}Moirai\\ (Large)\end{tabular}}  
    & 0-shot    & 31.30 & \underline{20.43}  & 10.20  & \underline{11.39}  & \underline{12.94}  & \underline{8.01}  \\
   & Full FT       & \underline{26.84}  & 23.69 & \underline{9.74}  & 12.30 & 16.09  & 8.44 \\
   & LoRA FT       & 35.21  & 25.39  & 11.18  & 11.66  & 16.70  & 8.23  \\  \hline

\multirow{3}{*}{\begin{tabular}[c]{@{}c@{}}Chronos\\ (Tiny)\end{tabular}}  
    & 0-shot    & 25.60  & 21.40  & 8.64  & 7.37  & 11.41  & 5.97  \\
   & Full FT       & 19.90  & 20.51  & 8.03  & 8.80  & 14.99 & 6.86  \\
   & LoRA FT       & \underline{19.79} & \underline{19.86} & \underline{7.90}  & \underline{7.22}  & \underline{11.17} & \underline{5.90}  \\ \hline
   
\multirow{3}{*}{\begin{tabular}[c]{@{}c@{}}Chronos\\ (Mini)\end{tabular}}  
    & 0-shot    & 25.21  & 20.44  & 8.52  & 7.44  & 11.20  & 5.97 \\
   & Full FT       & 20.50 & 21.80 & 8.19 & 10.46 & 17.45  & 7.50  \\
   & LoRA FT       & \underline{20.05}  & \underline{20.72}  & \underline{8.03}  & \underline{7.26}  & \underline{10.88} & \underline{5.90}  \\ \hline
   
\multirow{3}{*}{\begin{tabular}[c]{@{}c@{}}Chronos\\ (Small)\end{tabular}} 
    & 0-shot    & 25.04  & \underline{20.28}  & 8.49  & 7.19  & 11.01  & 5.92 \\
   & Full FT       & 20.93  & 20.95 & 8.23  & 10.04 & 16.53  & 7.37  \\
   & LoRA FT       & \underline{19.89}  & 20.44  & \underline{8.02}  & \textbf{7.08} & \underline{10.83}  & \textbf{5.88}  \\ \hline
   
\multirow{3}{*}{\begin{tabular}[c]{@{}c@{}}Chronos\\ (Base)\end{tabular}}  
    & 0-shot    & 25.70 & \underline{20.32}  & 8.53  & 7.33  & 11.02  & 5.96  \\
   & Full FT       & 20.80  & 21.09 & 8.20  & 10.15 & 16.82  & 7.37 \\
   & LoRA FT       & \underline{20.12}  & 21.06  & \underline{8.06}  & \underline{7.25}  & \underline{10.96}  & \underline{5.93}  \\  \hline
   
\multirow{3}{*}{\begin{tabular}[c]{@{}c@{}}Chronos\\ (Large)\end{tabular}} 
    & 0-shot    & 25.54  & \underline{19.75}  & 8.50  & 7.21  & 10.84 & 5.93  \\
   & Full FT       & 21.01  & 20.58  & 8.22  & 9.56  & 16.45  & 7.32  \\
   & LoRA FT       & \underline{20.00}  & 20.34  & \underline{8.06} & \underline{7.16}  & \underline{10.76} & \underline{5.91}  \\  \hline

\end{tabular}
\label{full}
\scriptsize *MSE values are scaled by 1e-4, and DTW values are scaled by 1e-3 for better comprehension.

\end{table}

\section{Experiment Setup}
\noindent{\textbf{Dataset}}:
For our experiments, we used the publicly available eICU Collaborative Research Database \cite{eicu} wherein we forecast the mean blood pressure (MeanBP) and heart rate (HR) vitals for patients diagnosed with sepsis or septic shock. Following \cite{spass1, spass2},  we clean and preprocess the data by imputing any missing values using the forward fill method. Thereafter, we extract 9 hours of both the vitals preceding diagnosis, with the initial 6-hour vitals serving as the context and the remaining 3 hours serving as the horizon window, which the model has to forecast. The vitals are sampled at intervals of 5 minutes, implying a context window of length 72 and a horizon window of length 36. Subsequently, we apply a low-pass filter to remove high-frequency noise in the time series data. We apply global min-max scaling to normalize the samples. Finally, we obtained 4020 samples, each for MeanBP and HR, from 1442 patients, which we divided into an 8:1:1 train, validation, and test split for fine-tuning and evaluating the models. \\

\noindent{\textbf{Implementation and Evaluation Details}}: For fine-tuning with LoRA, we follow the setup described in \cite{lora}, where all weight matrices in the attention module (Query, Key, Value, and Output) are adapted with low-rank matrices to assess LoRA's adaptability with minimal memory usage. The rank $r$ is set to 2, and the scaling factor $\alpha$ is set to 16. Given the different models, data types, and fine-tuning configurations (whether fine-tuning all model parameters or only the LoRA parameters), learning rates ranging from 1e-3 to 5e-5 are used to optimize performance using Adam as an optimizer. 
Aside from these changes, other hyperparameters remain consistent with those from the original model checkpoints. The implementation is done in PyTorch, and experiments are conducted on an NVIDIA A100 GPU. 
To ensure a comprehensive evaluation, we use mean squared error (MSE), dynamic time warping (DTW) distance, and mean average percentage error (MAPE) to quantify the differences between predicted forecasts and the ground truth. Given the probabilistic nature of the forecasts, the predicted value for each forecasted datapoint is taken as the median of 20 samples, which is then used to calculate the metrics. Each metric is also averaged over 10 forecasting runs to provide an accurate assessment and clearer understanding of model performance.

\begin{figure}[t]
  \centering
  \includegraphics[width=0.8\linewidth]{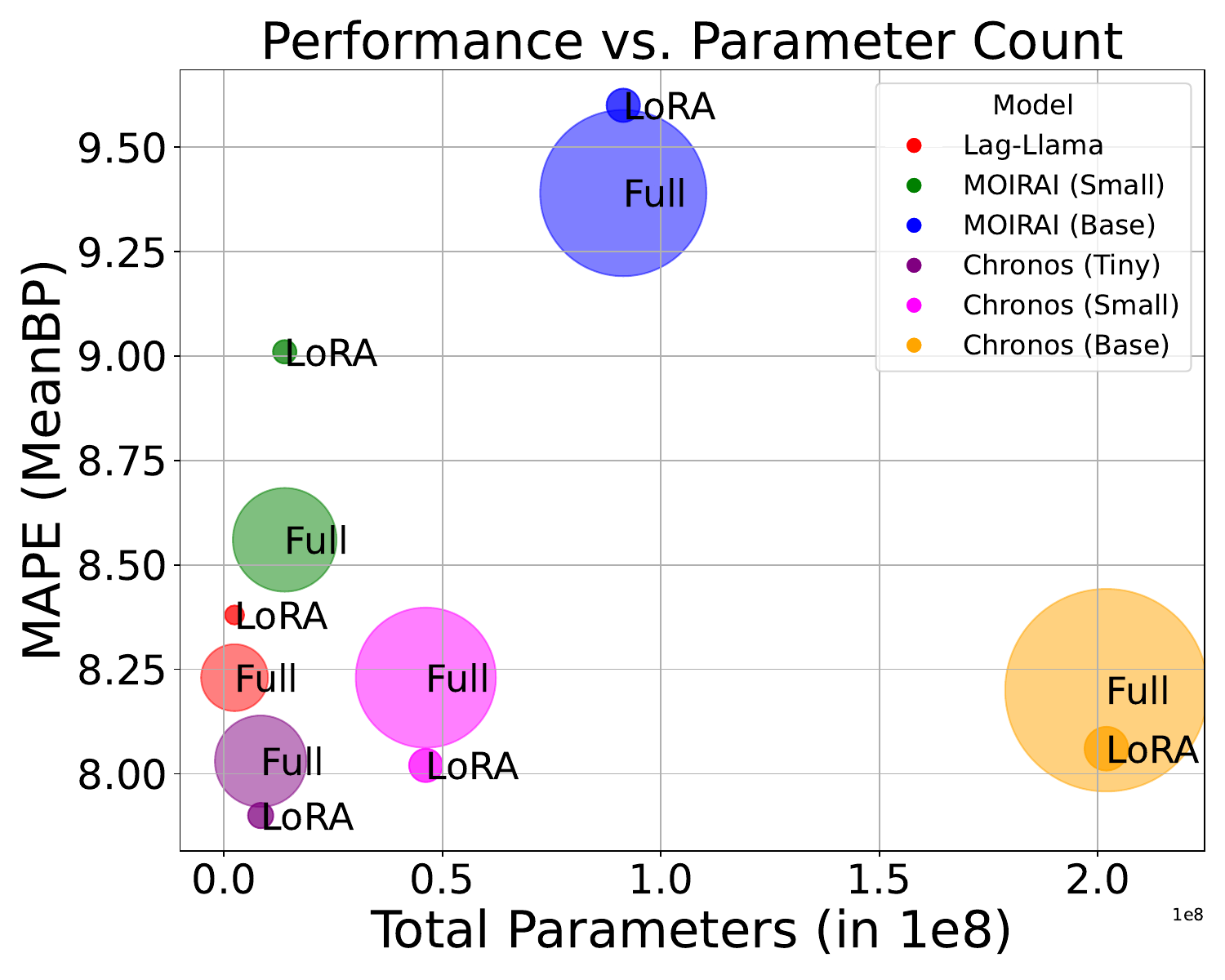}
  \caption{Performance across varying model parameter settings. Marker size indicates the number of fine-tuned parameters. Values are scaled appropriately for clarity.}
  \label{params}
\end{figure}

\section{Results and Discussion}
We present our comparative results in Table \ref{full}, which outlines the forecasting performance across different tuning methodologies: zero-shot (without fine-tuning), full fine-tuning of all original model weights, and selective fine-tuning using LoRA-specific adapter weight matrices. Our findings highlight that, among the evaluated TSFMs, Chronos consistently outperforms Lag-Llama and MOIRAI in zero-shot forecasting scenarios, as substantiated by prior studies \cite{chronos}. Specifically, full fine-tuning yields superior results in predicting MeanBP for the Lag-Llama model. Conversely, when employing LoRA-based fine-tuning, we observe enhanced performance metrics for HR forecasts compared to full fine-tuning. For MOIRAI, full fine-tuning delivered superior performance, closely followed by results from LoRA fine-tuning. However, in certain cases (HR - Base and Large, MeanBP - Large), full fine-tuning led to a performance decline. Unlike other TSFM candidates, MOIRAI processes time series by dividing them into patches and explicitly incorporates time-frequency information via distinct learnable weights for different frequencies. These weights, which form large parameter sets, significantly increase computational requirements when fine-tuning the entire model. We hypothesize that the decline in performance observed when fine-tuning larger models can be attributed to this parameter-heavy patch embedding and decoding layers (which are omitted in LoRA fine-tuning). 

\begin{figure}[t]
  \centering
  \includegraphics[width=0.8\linewidth]{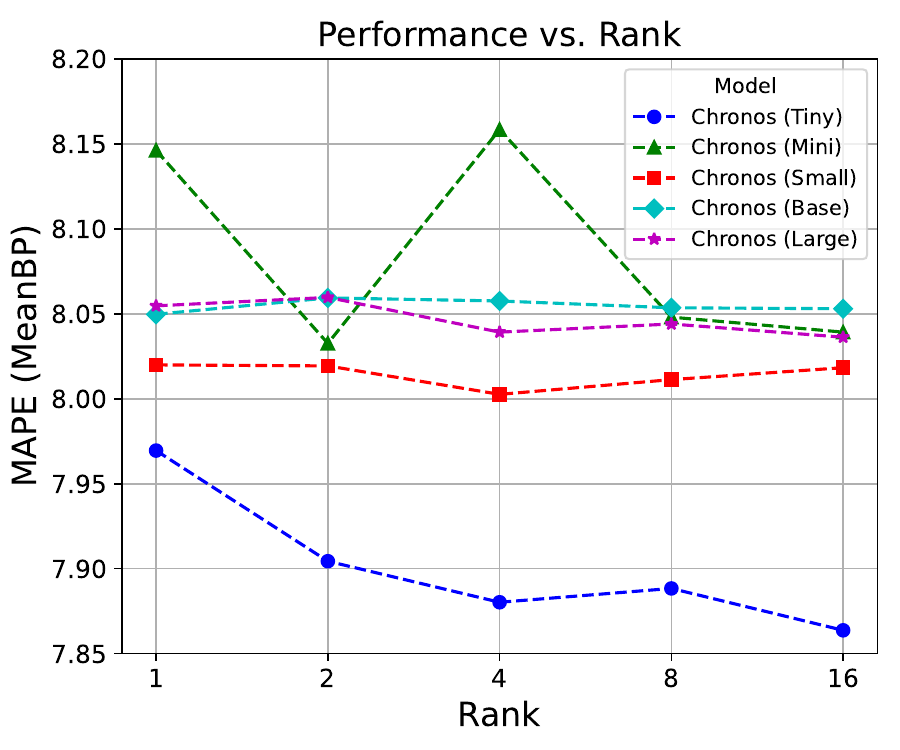}
  \caption{Performance across varying matrix ranks during LoRA fine-tuning for different models. }
  \label{rplot}
\end{figure}

For Chronos, full fine-tuning and LoRA fine-tuning significantly improved MeanBP forecasting performance, with LoRA fine-tuning consistently outperforming full fine-tuning. Although LoRA fine-tuning achieved better results in HR forecasting than the other two methods, the improvements were less pronounced compared to MeanBP. Consistent with Lag-Llama and MOIRAI, a decline in HR forecasting performance was observed with full fine-tuning. This may be attributed to the relatively lower variability of HR vital signs compared to MeanBP.

Additionally, we benchmark our results by comparing them against the 
SOTA approach for this task \cite{spass1}. However, it is worth noting that this comparison is not entirely equitable because the authors in \cite{spass1} trained multiple time series forecasting models from scratch exclusively on the 3,216 training MeanBP and HR time series samples. In contrast, TSFMs have processed millions (if not billions) of samples during pre-training and were fine-tuned using only a few thousand samples. We observe that some Chronos variants are fine-tuned with the LoRA approach SOTA performance levels. In particular, Chronos (Small) with LoRA fine-tuning surpasses the SOTA in forecasting HR vital signs across multiple metrics, while Chronos (Tiny) with LoRA fine-tuning closely approaches SOTA for MeanBP vital signs.

To better understand the impact of LoRA, we visualized the effect of fine-tuning varying numbers of parameters in Figure \ref{params}. The plot presents the MAPE values for MeanBP against the total parameter count, with marker sizes appropriately scaled to reflect the number of fine-tuned parameters. The total parameter count ranges from 2.42 million (2.41 million from original model weights and 16 thousand from LoRA weights) for Lag-Llama to 201.8 million (201.4 million from original model weights and 0.4 million from LoRA weights) for Chronos (Base). We observe that LoRA significantly reduces the number of parameters required for fine-tuning while achieving comparable, and often superior, performance compared to fully fine-tuned models. Notably, LoRA preserves FMs' generalization capabilities by retaining the original pre-trained weights, allowing multiple sets of LoRA weights to be trained for various tasks or data. This leads to the highly efficient use of TSFMs. To understand the impact of using different ranks $r$ to determine the LoRA weight matrices, we plotted the MeanBP MAPE values for varying $r$ in Figure \ref{rplot}. 
Our observations reveal that performance improvement typically plateaus after a certain $r$, indicating a LoRA adaptation threshold. Additionally, with increasing $r$, changes in MAPE values are more pronounced for smaller models, while larger models show minimal effect.

\section{Conclusion and Future Work}
This study comprehensively explored the efficiency and impact of LoRA-based fine-tuning across three leading TSFM architectures: Lag-Llama, MOIRAI, and Chronos. Our comparative analysis of zero-shot, full fine-tuning, and LoRA-based selective fine-tuning methodologies demonstrated that LoRA fine-tuning improves forecasting performance while significantly reducing computational demands compared to full fine-tuning. Furthermore, we showed that LoRA fine-tuning allows certain Chronos variants to match or exceed the SOTA in vital sign forecasting. Our ablation studies and evaluation of different LoRA matrix rank ($r$) revealed the delicate trade-offs between performance and trainable parameter count, highlighting the applicability and efficiency of LoRA fine-tuning in practical forecasting tasks. For future work, we plan to extend the low-rank adaptation to multivariate settings upon further advancement in the foundational model space and possibly explore combinations of multiple PEFT methods for improved performance.

\bibliographystyle{ACM-Reference-Format}
\bibliography{sample-base}

\end{document}